\documentclass[10pt,twocolumn,letterpaper]{article}

\usepackage{cvpr} 
\usepackage{multirow}
\usepackage{xcolor}

\definecolor{cvprblue}{rgb}{0.21,0.49,0.74}
\usepackage[pagebackref,breaklinks,colorlinks,allcolors=cvprblue]{hyperref}
\newcommand\blfootnote[1]{\begingroup\renewcommand\thefootnote{}\footnote{#1}\addtocounter{footnote}{-1}\endgroup}

\def\paperID{*****} 
\def\confName{CVPR}
\def\confYear{2026}

\title{EfficientSAM3: Progressive Hierarchical Distillation for Video Concept Segmentation from SAM1, 2, and 3}

\author{Chengxi Simon Zeng\thanks{Equal contribution} \quad Yuxuan Jiang\footnotemark[1] \quad Aaron Zhang \\
University of Bristol\\
{\tt\small \{simon.zeng, yuxuan.jiang, aaron.zhang\}@bristol.ac.uk}
}

\begin{document}
\maketitle
\blfootnote{\textcolor{red}{Coming Soon - This document details the experimental setup and training procedures for making SAM3 efficient. Implementation may evolve; results will be reported in a future revision.}}

\begin{abstract}
The Segment Anything Model 3 (SAM3) advances visual understanding with Promptable Concept Segmentation (PCS) across images and videos, but its unified architecture (shared vision backbone, DETR-style detector, dense-memory tracker) remains prohibitive for on-device use. We present EfficientSAM3, a family of efficient models built on \textbf{Progressive Hierarchical Distillation (PHD)} that transfers capability from SAM3 to lightweight students in three stages: (1) \emph{Encoder Distillation} aligns image features via prompt-in-the-loop training on SA-1B; (2) \emph{Temporal Memory Distillation} replaces dense memory with a compact Perceiver-based module trained on SA-V to compress and retrieve spatiotemporal features efficiently; and (3) \emph{End-to-End Fine-Tuning} refines the full pipeline on the official SAM3 PCS data to preserve concept-level performance. PHD yields a spectrum of student variants using RepViT, TinyViT, and EfficientViT backbones, enabling on-device concept segmentation and tracking while maintaining high fidelity to teacher behavior. We benchmark on popular VOS datasets, and compare with varies of releated work, achieing strong performance-efficiency trade-offs. 
\end{abstract}

\section{Introduction}
\label{sec:intro}
Foundation models for visual segmentation have rapidly evolved, fundamentally changing how machines perceive and interact with the visual world. The Segment Anything Model (SAM) series~\cite{sam1, sam2, sam3} has been at the forefront of this revolution. The initial SAM~\cite{sam1} introduced the paradigm of promptable zero-shot image segmentation, demonstrating an unprecedented ability to segment any object based on simple geometric prompts. Its successor, SAM2~\cite{sam2}, extended this capability to the temporal domain of video by incorporating a memory mechanism to track objects across frames. The latest iteration, SAM3~\cite{sam3}, represents another paradigm shift, moving from generic object segmentation to Promptable Concept Segmentation (PCS). SAM3 can detect, segment, and track all instances of a specific semantic concept provided via phrases or image exemplars, achieving a new level of semantic understanding.

This progression in capability has been accompanied by a corresponding increase in architectural complexity and computational cost. While SAM1's\footnote{We explicitly refer to the original Segment Anything Model~\cite{sam1} as SAM1 to distinguish it from its successors SAM2 and SAM3.} primary bottleneck was its massive ViT-H image encoder~\cite{sam1}, SAM2 introduced a second bottleneck in its memory attention module~\cite{edgetam}. SAM3 further complicates this with a unified architecture featuring a powerful shared vision backbone, a DETR-based detector with a decoupled presence head, and a memory-based video tracker~\cite{sam3}. These components, while enabling state-of-the-art performance on the new SA-Co benchmark~\cite{sam3}, render SAM3 even more computationally demanding than its predecessors, confining its use to server-side environments and precluding deployment in real-time, on-device applications such as augmented reality~\cite{sam_ar_song2024, sam_ar_liu2024}, robotics~\cite{sam_robotics_wang2023, sam_robotics_gupta2024}, medical image~\cite{zeng2025tuningvisionfoundationmodel, zeng2025agglomeratinglargevisionencoders} and interactive mobile tools~\cite{mobilesam_zhang2023, mobilesamv2_zhang2023}.

This computational barrier motivates our work. Following the successful symbiotic cycle of innovation where foundational model releases are followed by community-driven efficiency efforts~\cite{edgesam, efficientsam, edgetam}, we introduce \textbf{Progressive Hierarchical Distillation (PHD)} to create a family of efficient SAM3 variants for on-device deployment. PHD targets two primary bottlenecks in SAM3—(i) the shared vision backbone for feature extraction and (ii) the dense memory tracker for temporal consistency—and concludes with a unifying end-to-end refinement.

To address these, PHD proceeds in three stages. First, inspired by prompt-in-the-loop distillation in EdgeSAM~\cite{edgesam}, we distill the SAM3 encoder into efficient student backbones—RepViT~\cite{repvit}, TinyViT~\cite{tinyvit}, EfficientViT~\cite{efficientvit}. Second, following EdgeTAM/EfficientTAM~\cite{edgetam, efficienttam}, we replace SAM3's dense memory with a compact Perceiver~\cite{perceiver} that compresses temporal context into a small set of latent queries. Third, we jointly fine-tune the student encoder, Perceiver memory, and mask decoder end-to-end to preserve SAM3's PCS behavior.

\begin{figure*}[t]
  \centering
\includegraphics[width=\textwidth]{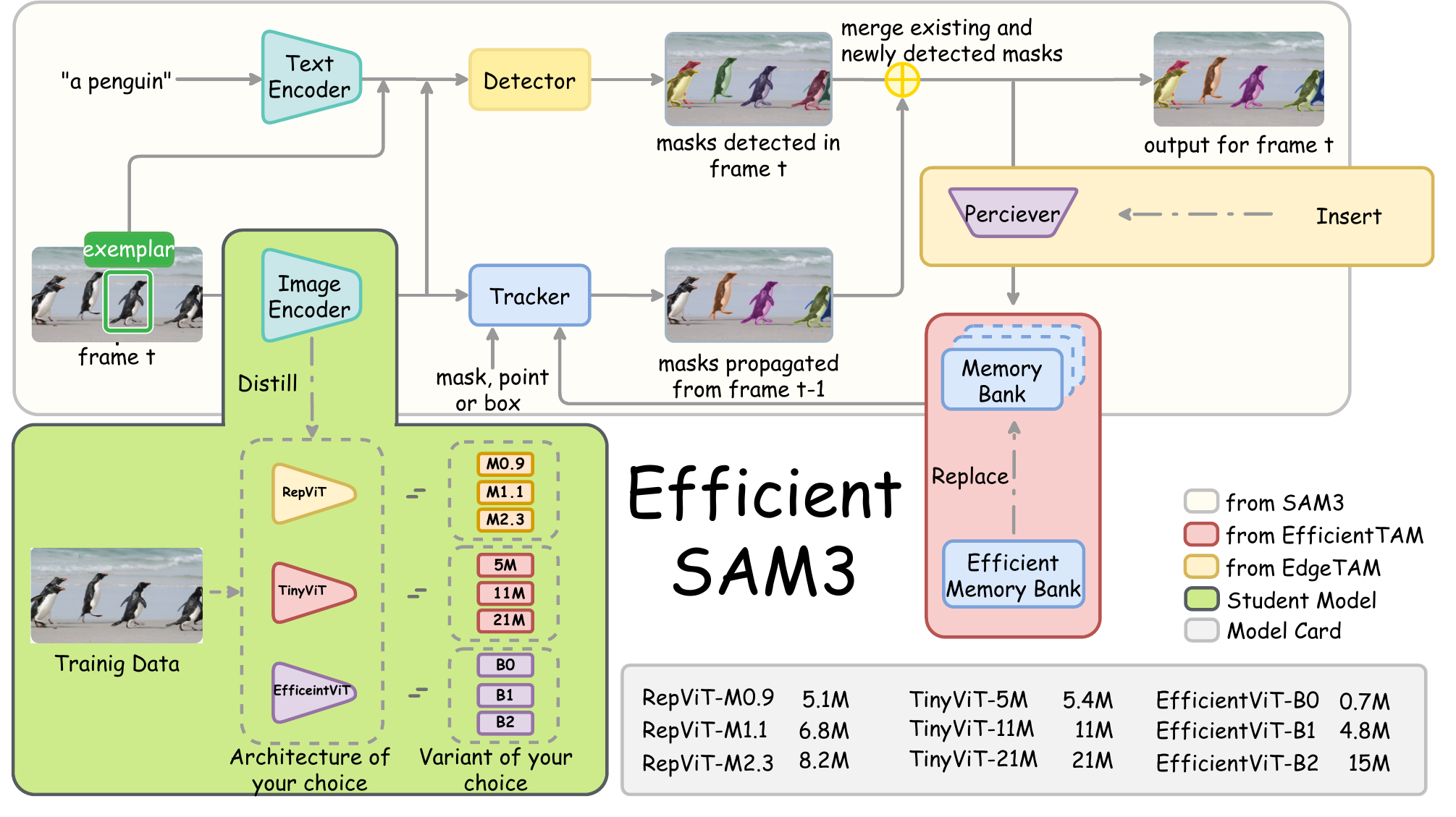}
  \caption{Overall architecture of EfficientSAM3: Three stages of Progressive Hierarchical Distillation (PHD) are shown—lightweight encoder distillation, compact Perceiver insertion and efficient memory bank replacement.}
  \label{fig:overall_arch}
\end{figure*}

Our main contributions are as follows:
\begin{itemize}
    \item We introduce EfficientSAM3, a \emph{three-stage} distillation recipe (encoder $\Rightarrow$ memory $\Rightarrow$ end-to-end) that enables on-device PCS while preserving teacher behavior.
    \item We build a \emph{model zoo} of nine EfficientSAM3-PHD students spanning RepViT, TinyViT, and EfficientViT, offering flexible accuracy-latency trade-offs for deployment.
    \item We compress temporal memory with a \emph{Perceiver}-based memory module, yielding efficient long-term reasoning and reduced memory-attention cost.
    \item We benchmark across \emph{SA-Co (PCS)}, \emph{SA-V/VOS} (COCO, LVOS, DAVIS17, MOSE, YTVOS19 and etc). 
\end{itemize}

\section{Related Work}
\label{sec:related_work}

\paragraph{Foundation Models for Segmentation.}
The Segment Anything Model (SAM) series~\cite{sam1, sam2, sam3} has established the paradigm of promptable segmentation. SAM1~\cite{sam1} introduced the core task for images using geometric prompts, powered by a massive ViT-H encoder trained on the extensive SA-1B dataset~\cite{sam1}. SAM2~\cite{sam2} extended this to videos by adding a memory mechanism for temporal tracking, defining the Promptable Visual Segmentation (PVS) task and introducing the SA-V dataset. Most recently, SAM3~\Cref{sec:sam3_preliminary} pivots to Promptable Concept Segmentation (PCS), where the model segments all instances of a semantic concept from noun phrases or exemplars, trained on the new SA-Co dataset. This shift towards semantic understanding, while powerful, increases architectural complexity, creating a pressing need for efficiency that our work directly addresses.

\paragraph{Efficient Image Segmentation Models.}
The computational cost of SAM1 spurred a wave of research into efficient variants. These efforts can be broadly categorized into backbone replacement and knowledge distillation~\cite{sam_survey}. Models like FastSAM~\cite{fastsam} and EdgeSAM~\cite{edgesam} replaced the ViT with efficient CNNs like YOLOv8~\cite{yolov8} and RepViT~\cite{repvit}, respectively. Others, like MobileSAM~\cite{mobilesam}, used a lightweight TinyViT~\cite{tinyvit}. Knowledge distillation proved crucial for retaining performance. While MobileSAM used a simple decoupled distillation, EdgeSAM~\cite{edgesam} introduced a more effective "prompt-in-the-loop" strategy that distills the entire interactive pipeline, better preserving the model's behavior. EfficientSAM~\cite{efficientsam} proposed SAMI, a masked image pre-training method to distill features from the teacher's intermediate layers. Our work builds on these principles, adapting the successful prompt-in-the-loop distillation of EdgeSAM to the more complex, concept-driven architecture of SAM3.

\paragraph{Efficient Video Segmentation Models.}
Optimizing video models like SAM2 presents a dual bottleneck: the image encoder and the memory attention module. EdgeTAM~\cite{edgetam} was the first to explicitly identify and solve the memory bottleneck. Its core contribution is the 2D Spatial Perceiver, which compresses dense memory features into a small set of latent queries before the costly attention operation, enabling a 22x speedup on mobile devices. EfficientTAM~\cite{efficienttam} also proposed optimizations for both the encoder and an efficient memory cross-attention mechanism. Our work adopts a similar philosophy to EdgeTAM, leveraging a Perceiver model to compress the memory bank in our EfficientSAM3 architecture, thereby tackling the temporal bottleneck which is crucial for efficient video concept tracking. This is complemented by a rich body of work in efficient Video Object Segmentation (VOS) that has explored memory mechanisms~\cite{xmem, stcn}, template matching~\cite{feelvos}, and transformer architectures~\cite{deva, trackformer, 11084428}.

\paragraph{Lightweight Vision Architectures.}
Our work leverages several state-of-the-art efficient backbones. \textbf{RepViT}~\cite{repvit} is a pure CNN architecture inspired by ViT design principles that uses structural re-parameterization to achieve exceptional on-device latency. \textbf{TinyViT}~\cite{tinyvit} is a small ViT designed for effective knowledge distillation from large teacher models. \textbf{EfficientViT}~\cite{efficientvit} introduces a multi-scale linear attention mechanism that reduces computational complexity from quadratic to linear, making it ideal for high-resolution tasks. By distilling SAM3 into these diverse backbones, we create a family of models that cater to various deployment constraints, from minimal parameter count to maximum throughput. Other relevant architectures include MobileNet~\cite{mobilenetv3} and ShuffleNet~\cite{shufflenetv2}, which have set the standard for efficient CNN design.

\paragraph{Knowledge Distillation and Model Compression.}
Knowledge distillation~\cite{hinton_distillation} is a cornerstone of our approach. Beyond its application in efficient SAMs, it has a rich history in model compression. Techniques range from matching output logits to intermediate features~\cite{fitnets} and attention maps~\cite{attention_distill}. Our prompt-in-the-loop methodology is a form of task-specific distillation, ensuring that the student model learns the final behavior required for concept segmentation. This is complemented by other compression techniques like quantization~\cite{quantization} and pruning~\cite{pruning}, which could be applied orthogonally to our method for further gains. \emph{Learning using privileged information (LUPI)} formalizes the idea of exposing a teacher to additional modalities or annotations available only at training time~\cite{vapnik2009lupi}. Our Stage~2 memory distillation fits this paradigm: the dense teacher memory acts as privileged information that supervises a compact Perceiver-based memory. \emph{Mutual distillation} eliminates the fixed teacher by allowing peers to learn from each other online~\cite{zhang2018dml, anil2018codistillation}, while \emph{self-distillation} and EMA teacher variants exploit iterative or temporal ensembling~\cite{furlanello2018bornagain, tarvainen2017meanteacher}. \emph{Contrastive representation distillation (CRD)} aligns student and teacher instance-level representations via a contrastive objective~\cite{tian2020crd}. \emph{Relational distillation} transfers higher-order structure—pairwise distances and local similarities—rather than absolute features~\cite{park2019rkd, tung2019spkd}. \emph{Teacher Assistant KD (TAKD)} introduces intermediate teachers to stabilize transfer when capacity gaps are large~\cite{mirzadeh2020takd}. Many of these ideas are orthogonal to PHD and can augment our three-stage pipeline without altering the deployment footprint.

\paragraph{Efficient Temporal Modeling.}
Our use of a Perceiver to compress the memory bank is part of a broader trend towards more efficient temporal modeling in video understanding. The Perceiver IO~\cite{perceiverio} demonstrated that a small set of latent queries can effectively process large, multimodal inputs by iteratively attending to them. This principle has been applied to various video tasks. Alternatives to transformer-based memory include state-space models like Mamba~\cite{mamba}, which offer linear-time sequence modeling and have shown promise in vision tasks~\cite{vim}. These represent promising future directions for further optimizing the memory mechanism in models like EfficientSAM3.

\section{Methodology}
\label{sec:method}

In this section, we first provide a formal preliminary of the Segment Anything Model 3 (SAM3). We then detail our proposed EfficientSAM3, outlining our prompt-in-the-loop distillation strategy and the integration of a Perceiver-based memory bank for efficient concept tracking.

\subsection{Preliminary: SAM3}
\label{sec:sam3_preliminary}

SAM3~\cite{sam3} introduces the task of Promptable Concept Segmentation (PCS). It is a unified model that takes a concept prompt \( P_c \) (a noun phrase, an image exemplar, or both) and an image \( I \) or video \( V = \{I_1, \dots, I_T\} \) as input, and outputs a set of instance masks \( \{M_1, \dots, M_N\} \) and corresponding identities for all objects matching the concept.

The architecture consists of a shared vision backbone \( E_{\text{vision}} \), an image-level detector \( D_{\text{det}} \), and a memory-based video tracker \( D_{\text{track}} \).

\paragraph{Image-Level Concept Detection.} For an image \( I_t \), the vision backbone extracts features \( F_t = E_{\text{vision}}(I_t) \). The detector, a DETR-based model~\cite{detr}, takes these features and the encoded concept prompt \( P_c \) to produce a set of object detections \( \mathcal{O}_t = \{ (B_i, S_i, M_i) \}_{i=1}^N \), where \( B_i, S_i, M_i \) are the bounding box, score, and mask for the \( i \)-th instance. A key innovation is the \textbf{presence head}, which first predicts a global presence score \( p(\text{present} | P_c, I_t) \). The final score for each detection is a product of this presence score and the localization score, decoupling recognition from localization.

\paragraph{Video Concept Tracking.} For video, SAM3 combines detection and tracking. The tracker propagates masklets (spatio-temporal masks) \( \mathcal{M}_{t-1} \) from the previous frame to the current frame \( t \), producing predictions \( \hat{\mathcal{M}}_t \). This propagation is conditioned on a memory bank \( \mathcal{B} \) containing historical appearance information. The memory bank is updated with features from past frames. The model then merges the propagated masks \( \hat{\mathcal{M}}_t \) with new detections \( \mathcal{O}_t \) from the detector on the current frame to produce the final output \( \mathcal{M}_t \). The memory update for an object \( o \) can be formulated as:
\begin{equation}
    \mathcal{B}_t^o = \text{Update}(\mathcal{B}_{t-1}^o, E_{\text{mem}}(F_t, M_t^o)),
\end{equation}
where \( E_{\text{mem}} \) is the memory encoder. The mask prediction for the next frame is conditioned on this memory:
\begin{equation}
    \hat{M}_{t+1}^o = D_{\text{track}}(E_{\text{vision}}(I_{t+1}), P_c, \mathcal{B}_t^o).
\end{equation}
The computational cost of this process is dominated by two components: the shared backbone \( E_{\text{vision}} \) and the attention mechanism within \( D_{\text{track}} \) that queries the dense memory bank \( \mathcal{B} \). Both EfficientTAM and EdgeTAM~\cite{efficienttam, edgetam} have identified memory attention as a critical bottleneck for on-device video segmentation, motivating our use of a compact Perceiver-based alternative.

\paragraph{PCS Prompts and Data Engine.} SAM3 formalizes \emph{Promptable Concept Segmentation (PCS)} with concept prompts defined as \emph{short noun phrases}, image exemplars, or both; the model returns instance masks and identities for \emph{all} objects matching the concept in images and videos. The PCS data engine scales beyond prior SA-1B efforts by curating diverse media domains and constructing \emph{millions of unique concept labels}, including systematically generated hard negatives to stress recognition. Ambiguities in open-vocabulary concepts (e.g., subjective attributes) are addressed throughout data design, metrics, and modeling, with interactive refinement to disambiguate outputs. The detector is \emph{identity-agnostic} while the tracker separates identities; both share a vision encoder. While SAM3 targets short noun phrases, more complex language can be handled by composing with an MLLM at inference time.

\subsection{Progressive Hierarchical Distillation (PHD)}
Our goal is to create a family of efficient models, EfficientSAM3-PHD, that retain the powerful PCS capabilities of SAM3 while being suitable for on-device deployment. We achieve this through Progressive Hierarchical Distillation (PHD): a staged approach that first distills the vision backbone, then compresses temporal memory, and finally refines end-to-end behavior.

\paragraph{Stage 1: Encoder Distillation (Image-Level Segmentation).}
Inspired by EdgeSAM~\cite{edgesam}, we employ prompt-in-the-loop distillation to train a lightweight student \( S \) to mimic the SAM3 teacher \( T \) at the image level on SA-1B~\cite{sam1}. Given an image \( I \) and concept prompt \( P_c \), the student and teacher backbones produce features \( F^S \) and \( F^T \). We align features with a projection head and supervise outputs via masked losses with bipartite matching:
\begin{equation}
  \mathcal{L}_{\text{feat}} = \big\| \text{Proj}_S(F^S) - F^T \big\|_2^2,
\end{equation}
\begin{equation}
  \mathcal{L}_{\text{mask}} = \sum_i \mathcal{L}_{\text{Dice}}(M_i^S, M_{\sigma(i)}^T) + \mathcal{L}_{\text{Focal}}(M_i^S, M_{\sigma(i)}^T).
\end{equation}
The total loss combines task supervision and distillation: \( \mathcal{L}_{\text{total}} = \mathcal{L}_{\text{task}} + \lambda_1 \mathcal{L}_{\text{feat}} + \lambda_2 \mathcal{L}_{\text{mask}} \). We distill into three efficient families—RepViT~\cite{repvit}, TinyViT~\cite{tinyvit}, EfficientViT~\cite{efficientvit}—each offering complementary efficiency-accuracy trade-offs.

\paragraph{Stage 2: Temporal Memory Distillation (Video Tracking).}
We replace the dense memory bank with a compact Perceiver module~\cite{perceiver}, following EdgeTAM~\cite{edgetam}. Given memory features $F_{\text{mem}} \in \mathbb{R}^{H \times W \times C}$, a standard Perceiver flattens spatial dimensions and compresses via cross-attention with $K$ learnable latents $Q_{\text{lat}} \in \mathbb{R}^{K \times C}$:
\begin{equation}
 F_{\text{comp}} = \text{softmax}\!\Big(\frac{(Q_{\text{lat}} W_Q)(F_{\text{flat}} W_K)^T}{\sqrt{d_k}}\Big) (F_{\text{flat}} W_V),
\end{equation}
where $F_{\text{flat}} = \text{Flatten}(F_{\text{mem}})$ and $F_{\text{comp}} \in \mathbb{R}^{K \times C}$. However, this destroys 2D structure critical for segmentation. We adopt EdgeTAM's \textbf{2D Spatial Perceiver}, which partitions latents into global (attending to $F_{\text{flat}}$) and patch-level (attending to local $H \times W$ regions) groups to preserve spatial layout. We train this module on SA-V~\cite{sam2} to match the teacher's memory-conditioned masks and scores.

\paragraph{Stage 3: End-to-End Fine-Tuning (Concept Segmentation).}
Finally, we jointly refine the distilled encoder, Perceiver memory, and mask decoder on the official SAM3 PCS data~\cite{sam3}. This preserves promptable concept behavior while stabilizing the student pipeline for deployment.

\begin{figure}[t]
  \centering
  \includegraphics[width=\linewidth]{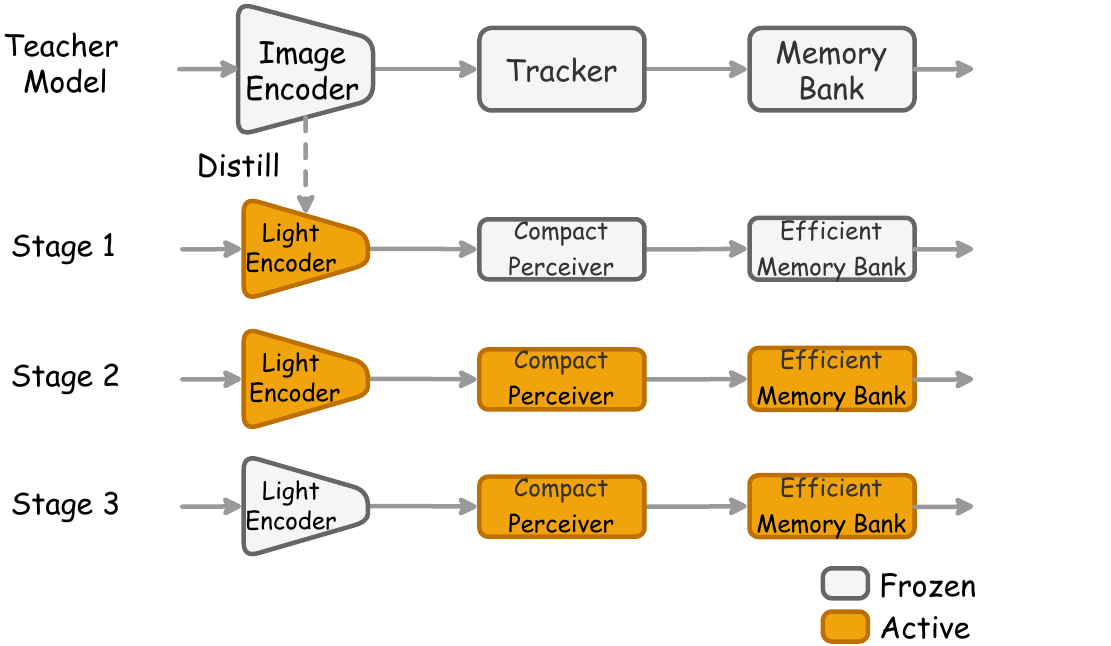}
  \caption{PHD training schedule showing the three-stage progression: Stage 1 distills the encoder on SA-1B, Stage 2 distills temporal memory on SA-V, and Stage 3 performs end-to-end fine-tuning on SA-Co. Each stage uses different loss compositions and freezes/trains different modules.}
  \label{fig:phd_schedule}
\end{figure}

\section{Training Setup and Procedures}
\label{sec:training}

This section describes the end-to-end training recipe for EfficientSAM3-PHD in sufficient detail to enable faithful reproduction. We focus entirely on data preparation, optimization, module freezing policies, prompt sampling, and engineering considerations; no quantitative results are reported here.

\subsection{Datasets and Preprocessing}
We organize training by stage and use datasets aligned with each objective:
\begin{itemize}
    \item \textbf{Stage 1 (image):} SA-1B~\cite{sam1}. We construct per-instance geometric prompts from masks: one GT box and the mask center point. When a dataset does not provide instance IDs, we treat connected components as instances.
    \item \textbf{Stage 2 (video):} SA-V~\cite{sam2}. We sample short clips (e.g., 4--8 frames) and expose the student to memory-conditioned decoding, using the SAM3 teacher to provide privileged dense memory supervision.
    \item \textbf{Stage 3 (PCS):} SA-Co~\cite{sam3}. We use short noun phrases and/or exemplar images as concept prompts. For videos, we pair concept prompts with clips sampled as in Stage~2.
\end{itemize}

Unless otherwise stated, images are resized such that the longer side is $\le$1024, preserving aspect ratio, then padded to multiples of 64. We apply standard augmentations used in promptable segmentation: random resize (short side in [640, 1024]), horizontal flip (p=0.5), color jitter (brightness/contrast/saturation/hue, small ranges), and random crop to $1024\times1024$ when needed. Inputs are normalized with ImageNet mean/std. Binary instance masks are rasterized at feature resolution for loss computation. Concept text is lowercased and tokenized by the SAM3 text encoder; exemplar images (if any) follow the same resize/normalize pipeline as the main image.

\subsection{Stage 1: Encoder Distillation with Prompts}
\label{sec:stage1}
We train a lightweight student encoder to mimic the SAM3 teacher under interactive usage, following the prompt-in-the-loop paradigm of EdgeSAM~\cite{edgesam} adapted to PCS. The student inherits the SAM-style prompt encoder and mask decoder interfaces; we add a projection/alignment head so that student features match the teacher decoder's expected shape (typically $(C{=}256, H{=}64, W{=}64)$ at $1024\times1024$ input). For CNN backbones (e.g., RepViT), we use a tiny FPN for resolution alignment as in \cite{edgesam}.

Given an image $I$ and a concept prompt $P_c$, we sample an \emph{initial} geometric prompt per instance (box or center point with equal probability). Teacher and student decode with the shared sparse prompt set, producing masks and scores. We then run one prompt-refinement loop ($M{=}1$ by default): compute the teacher–student disagreement region and sample a corrective point (positive from false negatives or negative from false positives), append it to the prompt set, and decode again. Losses are accumulated across the initial and refined prompts.

The optimization objective combines feature and mask distillation (see \Cref{sec:method}): $\mathcal{L}_{\text{feat}}$ aligns encoder features via an MSE on the projected student feature, and $\mathcal{L}_{\text{mask}}$ supervises binary mask logits using Dice+Focal losses with bipartite matching across the set of predicted masks. Unless otherwise noted, gradients flow through the student encoder and decoder; the teacher is frozen. We do not backpropagate through the teacher-generated target masks.

\noindent\textbf{Initialization and architecture.} Student encoders are initialized from ImageNet classification checkpoints when available (RepViT/TinyViT/EfficientViT). The student decoder is initialized from the SAM-family mask decoder weights to preserve interface behavior. We keep the text/exemplar encoders frozen in Stage~1 to decouple PCS language grounding from low-level feature alignment.

\noindent\textbf{Hyperparameters.} We use AdamW with cosine decay, base LR $1\times10^{-4}$, weight decay 0.05, and linear warmup for the first 1k steps. Batch size is 64 images (with gradient accumulation if limited by memory). We sample up to 16 instances per image to avoid VRAM overflow, following \cite{edgesam}. Mixed precision (AMP) is enabled. To lower overhead, we cache teacher encoder features to disk, as in \cite{edgesam}, and only recompute teacher decoder outputs when the prompt set changes.

\paragraph{Student Variants (Model Zoo).}
We instantiate nine students spanning three families and sizes; these differ only in encoder capacity and the alignment head, while the decoder and training losses remain identical:
\begin{table}[ht]
\centering
\caption{EfficientSAM3-PHD student variants used throughout training. Parameter counts are approximate.}
\label{tab:model_zoo}
\resizebox{.9\linewidth}{!}{%
\begin{tabular}{l|l|c}
\toprule
Model Name & Backbone & Parameters \\
\midrule
ES-RV-S & RepViT-M0.9 & 5.1M \\
ES-RV-M & RepViT-M1.1 & 6.8M \\
ES-RV-L & RepViT-M2.3 & 8.2M \\
\midrule
ES-TV-S & TinyViT-5M & 5.4M \\
ES-TV-M & TinyViT-11M & 11M \\
ES-TV-L & TinyViT-21M & 21M \\
\midrule
ES-EV-S & EfficientViT-B0 & 0.7M \\
ES-EV-M & EfficientViT-B1 & 4.8M \\
ES-EV-L & EfficientViT-B2 & 15M \\
\bottomrule
\end{tabular}}
\end{table}

\subsection{Stage 2: Temporal Memory Distillation}
\label{sec:stage2}
We replace the dense memory bank with a compact Perceiver~\cite{perceiver} trained to emulate the teacher's memory-conditioned decoding, following the 2D Spatial Perceiver design of EdgeTAM~\cite{edgetam}. The Perceiver compresses $F_{\text{mem}}\in\mathbb{R}^{H\times W\times C}$ from past frames into $K$ latent vectors. To preserve spatial structure, we partition latents into global and local groups that attend to the full feature map and to local windows, respectively. The compressed memory is queried by the tracking head to decode masks on the current frame conditioned on the concept prompt.

\noindent\textbf{Training clips and memory update.} We sample clips of $T\in[4,8]$ frames with stride 1. At each step $t$, the student updates its memory with $(F_t, M_t)$ and predicts masks for $t{+}1$; the teacher runs with its dense memory to produce supervision. The loss matches teacher masks and scores at each step (Dice+Focal for masks; BCE for presence/track scores when available). We optionally include a feature matching term from the teacher's memory readout to stabilize early training.

\noindent\textbf{Hyperparameters.} Unless noted, the student encoder remains frozen in Stage~2; we train the Perceiver memory and the tracking head. We use AdamW with base LR $5\times10^{-5}$, cosine decay, and warmup. Batch size is 16 clips; gradient checkpointing is enabled for the memory to reduce peak activation memory. We set $K{=}128$ latents by default. Temporal windows are randomly cropped to improve robustness to occlusion and reappearance.

\subsection{Stage 3: End-to-End PCS Fine-Tuning}
\label{sec:stage3}
Finally, we jointly refine the distilled encoder/Perceiver/decoder stack on SA-Co with PCS prompts (short noun phrases and/or exemplars). To preserve the distilled visual features, we freeze the student encoder by default and fine-tune the Perceiver memory and the mask decoder with a reduced LR. We form batches that mix images and short video clips (if available) to stabilize concept grounding across modalities.

\noindent\textbf{Losses and sampling.} The objective extends Stage~1/2 losses with concept-aware terms: we supervise the presence head and instance selection under concept prompts, and keep Dice+Focal for masks. Negative prompts and hard negatives (confusable concepts) are sampled to encourage precise recognition. For exemplar prompts, we randomly pick one or two exemplars per concept.

\noindent\textbf{Hyperparameters.} AdamW, base LR $2\times10^{-5}$ for trainable modules, cosine decay, and small warmup. Batch size 32 images or 8 clips. We keep AMP on, apply gradient clipping at 1.0, and use an exponential moving average (EMA) of student weights for stability.

\subsection{Engineering and Reproducibility Notes}
\begin{itemize}
    \item \textbf{Precision and memory.} All stages use PyTorch AMP (float16/bfloat16 depending on hardware). Gradient checkpointing is enabled for decoder cross-attention blocks and the Perceiver.
    \item \textbf{Freezing policy.} Teacher is always frozen. Stage~1 trains encoder; Stage~2 trains Perceiver+tracking head with encoder; Stage~3 trains Perceiver+decoder with encoder frozen (encoder unfreeze is optional for larger students).
    \item \textbf{Optimization.} Weight decay 0.05 (encoder) / 0.01 (decoder/memory); cosine LR with 5--10\% warmup; dropout disabled except in the Perceiver (p=0.1).
    \item \textbf{Prompt sampling.} One refinement loop ($M{=}1$) per instance is the default. We sample up to 16 instances per image for Stage~1 and up to 8 tracks per clip for Stage~2.
    \item \textbf{Teacher caching.} Following \cite{edgesam}, we cache teacher encoder features to reduce IO/compute; decoder outputs are recomputed only when prompts change.
    \item \textbf{Hardware.} Unless otherwise stated, training uses 8$\times$V100 with gradient accumulation to match the effective batch sizes above. Random seed is fixed to 42.
\end{itemize}

\subsection{Planned Evaluation Protocol (No Results yet)}
For completeness, we outline the metrics we will report in a results-bearing version: for images with geometric prompts (COCO/LVIS), mIoU; for videos (DAVIS/MOSE/SA-V/YTVOS), J\&F and $\mathcal{G}$ as appropriate; for PCS (SA-Co/VEval), CGF1 / IL MCC / pmF1 for images and CGF1 / PHOTA for videos. 

\section{Conclusion}
We introduced EfficientSAM3-PHD, a staged recipe to make SAM3 practical on-device while preserving its promptable concept capabilities. PHD progresses from encoder distillation (SA-1B), to temporal memory distillation with a Perceiver (SA-V), and finally to end-to-end PCS refinement (SA-Co). Across nine student variants spanning RepViT, TinyViT, and EfficientViT, PHD provides a flexible accuracy–latency trade-off. A subsequent version will report comprehensive quantitative results; this version focuses on training details for faithful reproduction.

\clearpage
{
    \small
    \bibliographystyle{ieeenat_fullname}
    \bibliography{main}
}

\end{document}